\documentclass[conference]{IEEEtran}
\IEEEoverridecommandlockouts
\usepackage{cite}
\usepackage{amsmath,amssymb,amsfonts}
\usepackage{algorithmic}
\usepackage{graphicx}
\usepackage{textcomp}
\usepackage{xcolor}
\usepackage{booktabs}
\def\BibTeX{{\rm B\kern-.05em{\sc i\kern-.025em b}\kern-.08em
    T\kern-.1667em\lower.7ex\hbox{E}\kern-.125emX}}
\begin{document}

\title{Variational Autoencoder for Calibration: A New Approach\\

\thanks{This work has been supported with funding from Sentech Soc Ltd, South Africa.}

}

\author{\IEEEauthorblockN{Travis Barrett}
\IEEEauthorblockA{\textit{Department of Electrical Engineering} \\
\textit{University of Cape Town}\\
Cape Town, South Africa \\
brrtra005@myuct.ac.za}
\and
\IEEEauthorblockN{Amit Kumar Mishra}
\IEEEauthorblockA{\textit{National Spectrum Centre} \\
\textit{Aberystwyth University}\\
Aberystwyth, UK \\
akmishra@ieee.org}

\and
\IEEEauthorblockN{Joyce Mwangama}
\IEEEauthorblockA{\textit{Department of Electrical Engineering} \\
\textit{University of Cape Town}\\
Cape Town, South Africa \\
joyce.mwangama@uct.ac.za}

}

\maketitle

\begin{abstract}
\textcolor{black}{In this paper we present a new implementation of a Variational Autoencoder (VAE) for the calibration of sensors. We propose that the VAE can be used to calibrate sensor data by training the latent space as a calibration output. We discuss this new approach and show a proof-of-concept using an existing multi-sensor gas dataset. We show the performance of the proposed calibration VAE and found that it was capable of performing as calibration model while performing as an autoencoder simultaneously. Additionally, these models have shown that they are capable of creating statistically similar outputs from both the calibration output as well as the reconstruction output to their respective truth data. We then discuss the methods of future testing and planned expansion of this work.}


\end{abstract}

%

\section{Introduction}

The use of low-cost sensors, or any sensors for that matter, requires a consistent management of sensor drift and calibration to achieve accurate measurements. This is particularly important when using low-cost sensors as they suffer from a range of challenges depending on the different sensor technologies. For air quality sensors, this is especially common as these devices are influenced by \textcolor{black}{various factors}. Cross-sensitivities, environmental factors and sensor degradation cause these devices to drift and become inaccurate over time \cite{concas2021low, piedrahita2014next}. This makes calibration critically important for optimal sensor performance. As stated by Liang and Daniels \cite{liang2022influences}, failure to invest in calibration may leave large uncertainties in retrieving reliable low-cost sensor data that further hinders its broader applications.

Machine Learning (ML) as a method for calibration has been widely discussed in open literature. Implementations of ML have been used to calibrate data in weather measurement stations for low-cost temperature sensors\cite{yamamoto2017machine}, for low-cost particulate matter sensors \cite{chen2018calibration, wang2017deep}, and many other implementation such as increasing the speed of calibration of Metal Oxide Semiconductor (MOS) sensors using transfer learning as shown by Robin \textcolor{black}{et al.} \cite{robin2022deep}. In the work done by Mu and Chen \cite{mu2022developing}, they show a calibration model building method based on semi-supervised learning is presented for in situ measurements of process components based on near-infrared (NIR) spectra where there are insufficient labeled samples. This use of a Conditional Variational Autoencoder (CVAE) allows for the generation of NIR spectra to increase the size of training data and when combined with the S2-LN calibration model, achieve a higher labeling accuracy than the other compared methods.

\textcolor{black}{Variational Autoencoders have been shown to function as calibration models as well as showing promise for drift detection. In the work done by Hossain et al. \cite{hossain2024enhancing}, a Variational Autoencoder (VAE) is presented for performing drift compensation and error detection. This implementation allows for processing data in parallel to enabling the Variational Autoencoder to compare the data to the distribution and correct the output accordingly. We also see the VAE used in the process of developing soft sensors, as seen in the works of Dai et al. \cite{dai2023latent} and Shen et al. \cite{shen2022predictive}. These works make use of the ability of the VAE to deal with non-linearity. Mihailescu \cite{mihailescu2021weakly} showed the use of a weakly supervised VAE for domain adaptation. It is clear that the VAE model has promising properties for calibration, error detection and domain adaptation, shown in current literature.}

In this paper we present a new use for the Variational Autoencoder (VAE) architecture as a method for sensor calibration. We propose that the VAE can be used to calibrate sensor data by training the latent space as a calibration output. This allows for the facilitation of calibration of sensor data while retaining the autoencoding functionality. In this paper we show a proof of concept for this approach to sensor calibration with the use of the dataset presented in De Vito \textcolor{black}{et al.} \cite{de2008field}. As shown in our previous work \cite{barrett2023agile}, Machine Learning (ML) could be a viable solution to sensor calibration. The intention of this work with new data and new model types is to find different approaches to this problem of sensor calibration with machine learning. Our contributions can be summarized as follows:
\begin{itemize}
    \item Present new VAE for calibration.
    \item Use the latent space of the VAE as a calibration output.
    \item \textcolor{black}{Demonstrate the model on an existing dataset.}\\
\end{itemize}

The remainder of the paper is organized in the following way. We will discuss the dataset used for \textcolor{black}{the} experimentation proposed in this paper. We will then describe the VAE and \textcolor{black}{the} proposed VAE for calibration. Following this we will present the methodology of experimentation used in this body of work. The results of the experimentation set out in the preceding section will then be shown. A discussion into the results will be presented and finally conclusions will be drawn about the work presented. With this conclusion, improvements to this work and future work will be presented.

\section{Dataset}
This section describes the dataset used for the work in this paper. \textcolor{black}{It gives the background information of the dataset and then describes the content of the dataset as described by the creators.}

\subsection{Information}
Created by De Vito \textcolor{black}{et al.} \cite{de2008field}, the dataset contains the responses of a multi-sensor gas sensing system deployed in an Italian city. This dataset includes these responses captured in the form of hourly averages as well as the recorded gas concentrations from a reference analyzer. This dataset has been used by many publications around sensor calibration and low-cost sensor performance. Some of these papers include further work by De Vito \textcolor{black}{et al.} \cite{de2009co} as well as the work by Yu \textcolor{black}{et al.} \cite{yu2020deep}. As this is a extensively used dataset it was a natural choice for our testing.

\subsection{Collected Data}
The dataset is described as having 9358 instances of hourly averaged responses from an array of 5 metal oxide chemical sensors (MOS). These readings were taken in the field in a polluted area of an Italian city at road-level. The data was collected with ground truth hourly averaged concentrations for CO, Non-methane Hydrocarbons, Benzene, Total Nitrogen Oxides (NOx) and Nitrogen Dioxide (NO$_2$)  and were provided by a co-located reference certified analyzer. This data recording took place from March 2004 to February 2005. Evidence of cross-sensitivities as well as both concept and sensor drift are present as described in the work done by De Vito \textcolor{black}{et al.} \cite{de2008field}.

\section{Variational Autoencoder}
This section discusses the Variational Autoencoder as well as the proposed calibration from latent space Variational Autoencoder.

\subsection{Variational Autoencoder}
The Variational Autoencoder is an autoencoder machine learning model. It is characterized as having two distinct networks in the model\cite{girin2020dynamical}. The first is the encoder network which takes the input data and maps it to a lower-dimensional latent space. The second being the decoder, which maps the lower-dimensional latent space back to the input data. The latent space of the Variational Autoencoder is a probabilistic latent space meaning that encoded inputs are mapped to a probabilistic distribution rather than a single point. A key component of the variational encoder is that in order to have the probabilistic latent space be backpropagated, stochastic backpropagation is used \cite{bengio2013representation}. This allows for the random noise to be separated from the statistical properties and allows the entire model to be backpropagated. The model may also be used as a generative model by sampling the latent space and decoding these with the decoder. This \textcolor{black}{enables the generation of data that was not originally present in the training data as a blend of properties that are related to the sampled area of the latent space.}

\subsection{New Proposed Variational Autoencoder}

\begin{figure*}
    \centering
    \includegraphics[width=0.9\linewidth]{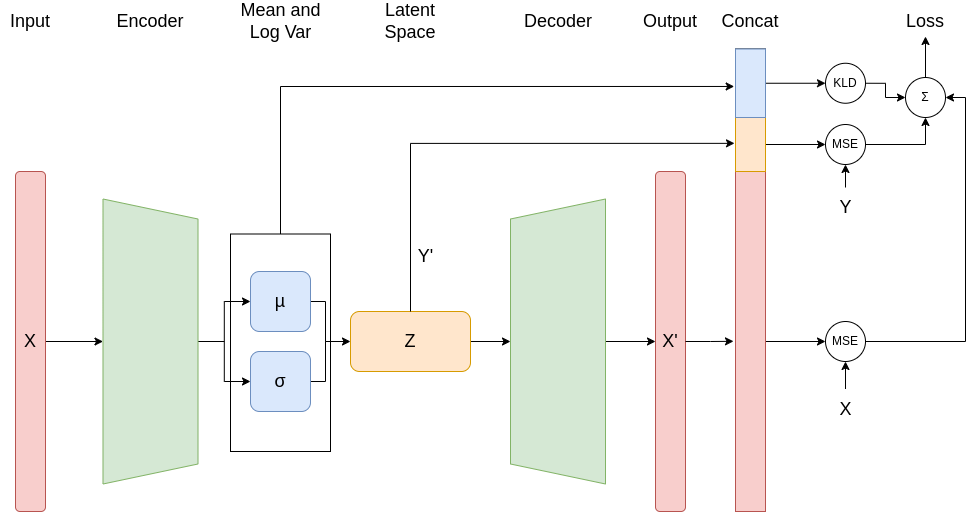}
    \caption{Proposed VAE Calibration Model Architecture where $X\approx X'$ and $Y\approx Y'$. The input to the model is represented by $X$ and $X'$ is the reconstruction, while $Y$ represents the truth value and $Y'$ represents the prediction.}
    \label{fig:Vae_prop}
\end{figure*}

A diagram of the proposed VAE can be seen in Figure \ref{fig:Vae_prop}. Using the variational encoder architecture, we seek to use the latent space sampling $Z$ as the output for calibration ($Y'$). This forces the latent space to fit the calibrated data characteristics and forces the encoder to be a predictor and the decoder as an expected input while retaining the reconstructive property of the original VAE. In Figure \ref{fig:Vae_prop}, the input to the model ($X$) can be seen. The model encodes $X$ down to the 1-dimensional (1D) latent space representation. From here, the model predicts $Y$, the truth data provided by a reference, through the latent space output $Y'$. The model decodes the latent space to reconstruct the input data $X$, producing $X'$. 

We have chosen to use this constrained form of latent space training a means of capturing the sensor properties in the VAE architecture. \textcolor{black}{In a traditional VAE the properties of the system are captured and represented in the latent space of the model through training the output of the model to approximate the input. We intend to use this property with the constraint on the latent space simultaneously during training.} We believe that in doing so the relationship of the sensor to the calibrated data at the time of training is captured \cite{mishra2023propagation}. First we are to determine if the calibration constrained latent space can successfully be implemented as a calibrator and a VAE simultaneously. In future we aim to explore the properties of the captured model weightings and biases as the window of time that the model is trained over. It is our hope that through the analysis of the changes of the weightings and biases that we may be able to compensate for \textcolor{black}{sensor} drift for correcting the models with respect to the differences between them.

\section{Methodology}

\textcolor{black}{As done previously in our work \cite{barrett2024statistical}, a set of metrics will to be used to give insight into the performance of the model.} These metrics were chosen because they give insight into the different components of the data. As before, we aim to quantify the performance with regard to the absolute values as well as the statistical properties of the data. The metrics we will use for comparison are Mean Absolute Error (MAE), Percentage Accuracy, R$^2$, as well as the Kullback-Leibler Divergence \textcolor{black}{(KLD)}, seen in Equation \ref{eq:KLDiv}, and Jensen-Shannon Divergence, seen in Equation \ref{eq:JSDiv}. 

\begin{equation}
    MAE = (\frac{1}{n})\sum_{i=1}^{n}\left | \hat{y}_{i} - y_{i} \right |
\label{eq:MAE}
\end{equation}

\begin{equation}
    Accuracy(\%) = 100 - \frac{1}{n}\sum_{i=1}^{n}(100\frac{\left | \hat{y}_{i} - y_{i} \right |}{y_{i}})
\label{eq:Acc}
\end{equation}

\begin{equation}
    \textcolor{black}{R^2 = 1 - \frac{\sum_{i=1}^{n}(y_{i}-\hat{y}_{i})^{2}}{\sum_{i=1}^{n}(y_{i}-\Bar{y})^{2}}}
\label{eq:r_squared}
\end{equation}

\begin{equation}
    KL(\hat{y} || y) = \sum_{c=1}^{M}\hat{y}_c \log{\frac{\hat{y}_c}{y_c}}
\label{eq:KLDiv}
\end{equation}

\begin{equation}
    JS(\hat{y} || y) = \frac{1}{2}(KL(y||\frac{y+\hat{y}}{2}) + KL(\hat{y}||\frac{y+\hat{y}}{2})) 
\label{eq:JSDiv}
\end{equation}
 Where:\\ $y$ is the truth value\\$\hat{y}$ is the predicted value\\$\Bar{y}$ is the mean value\\

Each model will be trained, and then the performance of each of the outputs will be evaluated using the metrics we have selected. This will allow for the comparison of the models and to determine the performance of this new proposed VAE implementation for this dataset. 

\subsection{Proposed VAE Model Properties}
\textcolor{black}{The proposed model, referring to the architecture visible in Figure \ref{fig:Vae_prop}, is to have 4 input nodes PT08.S1(CO), PT08.S2(NMHC), PT08.S3(NOx), and PT08.S4(NO$_2$). These 4 input nodes form $X$, the input data. It is then encoded into a 1D latent space (the calibration output $Y'$). This is then decoded back to 4 output nodes which are the reconstruction of the input data ($X'$). This decoded output is concatenated with the outputs of the $\mu$ and $\sigma$ nodes, as well as the calibration output. This is done to simplify the training process, as all the outputs can be passed to the loss function as a single vector. The truth data used for calibration, $Y$, is given by the chosen reference sensor of the chosen target pollutant. This means that for each reference sensor, CO(GT) as well as the other GT sensors in Figure \ref{fig:Correlation} are the truth data for their respective model.}

\subsubsection{\textcolor{black}{Implementation}}


\textcolor{black}{The VAE models were implemented in Google Colab using TensorFlow and Keras. This model is trained using a custom loss function as shown in Figure \ref{fig:Vae_prop}, which can be described by the following Equation:
\begin{multline*}
    Loss = \alpha \cdot  MSE(X, X') + \beta \cdot MSE(Y, Y')\\+ \gamma \cdot KLD(\mu , \sigma)
\end{multline*}
Where: $\alpha = 1$ and $\beta = 1$ and $\gamma = 0$\\
}

\subsubsection{\textcolor{black}{Hyper-Parameters}}
\textcolor{black}{The nodes used will be standard dense nodes using a sigmoid activation function to produce continuous values throughout the model. With the input and output dimension of the VAE being 4, the encoder and decoder layers are comprised of 4 nodes. Both $\mu$ and $\sigma$ layers are connected to the encoder and are single nodes. $Z$ is computed using a Lambda layer that takes the encoder outputs along with the $\mu$ and $\sigma$ outputs, implementing the standard normally distributed sampling function typically used in VAEs. The outputs from the decoder, the $\mu$, $\sigma$ and $Z$ layers are concatenated as the final model output. The model is trained with the Adam optimizer and the custom loss function shown before.}

\subsection{Data Preparation}

\begin{figure}
    \centering
    \includegraphics[width=\linewidth]{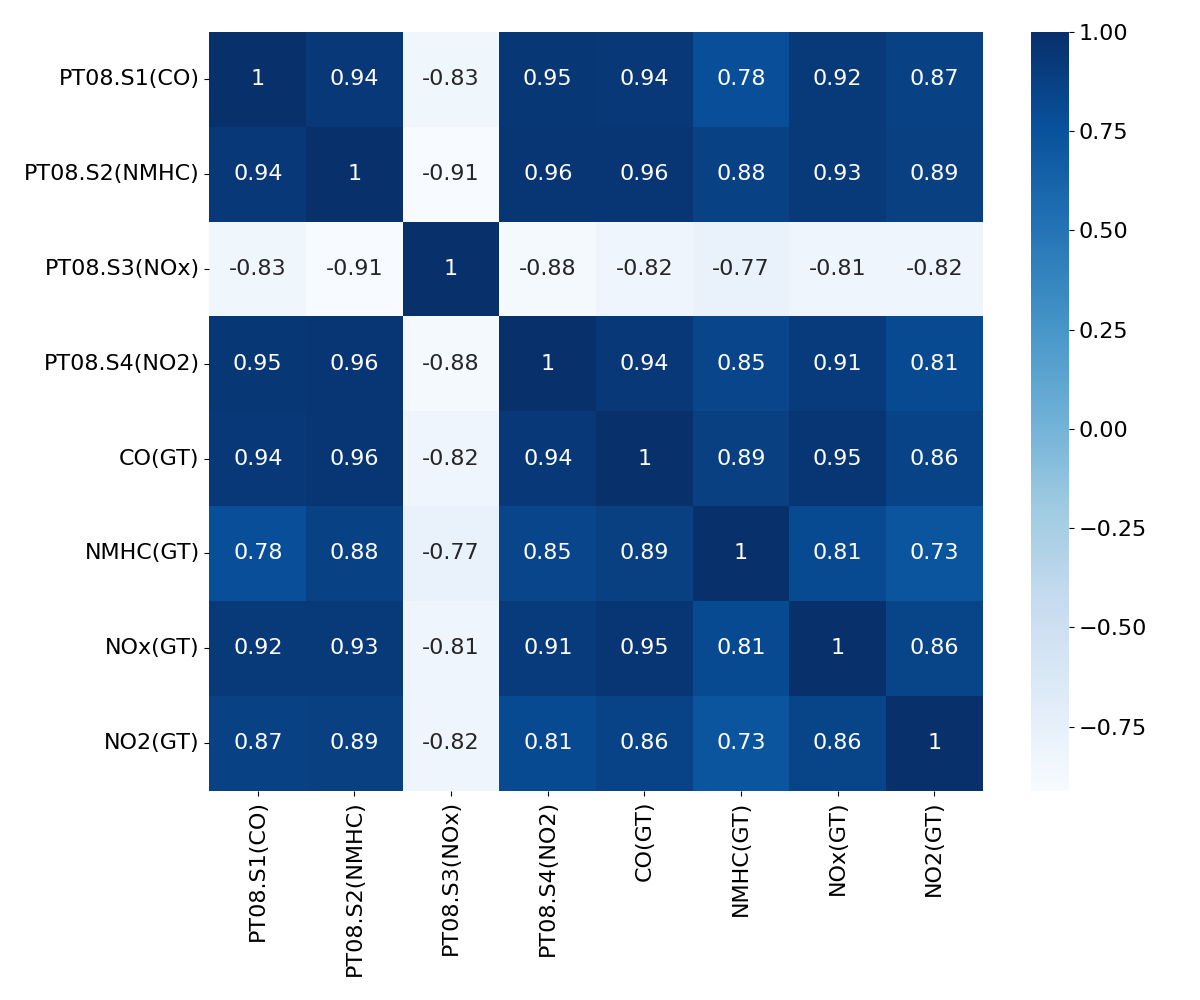}
    \caption{Correlation Coefficient Heatmap of Matched Test Dataset}
    \label{fig:Correlation}
\end{figure}

The dataset was prepared for our testing in a specific way. Firstly, we needed complete entries where all the sensors of interest were reporting and had a reference value to compare. This resulted in 827 complete readings for our testing. Each sensor's readings were then normalized using the min-max normalization method, seen in Equation \ref{eq:min-max_norm}, to scale and bias each sensor's data individually. The sensors chosen were all the MOS sensors in the dataset that showed strong correlation, or high cross-sensitivity. The correlation coefficients of the sensors and their matched ground truth sensors can be seen in Figure \ref{fig:Correlation}.

\begin{equation}
    x' = \frac{x - \min(x)}{\max(x) - \min(x)}
\label{eq:min-max_norm}
\end{equation}

\subsection{Training}
\textcolor{black}{In order to train the model, a custom error function was created to allow for the training of the model.} This was done by changing the standard output of the VAE to include the values from the latent space as well as the mean and log var nodes. This concatenation of nodes is shown in Figure \ref{fig:Vae_prop}. Specifically, in this test case, the $Z$ value from the sampling function of the latent space is the value we are training to be the calibrated output. This is done \textcolor{black}{using the} Mean Squared Error (MSE) between the predicted value and the truth data for the chosen sensor. Similarly the reconstruction error is calculated as the MSE of the reconstructed input data. It is important to note that the \textcolor{black}{Kullback-Leibler divergence loss} was not used in training of these models as it reduced the performance when using Gaussian distributions. Further investigation into the underlying data may allow for statistics-based loss functions in a future work. It is also worth noting that we are reducing the latent space to one dense node and, therefore, a singular calibrated value. 

The models were trained on the first 300 values of the dataset, with the data randomized into batches of 16. This was done as to see the performance of the model on the seen and unseen data due to the relatively small size of the chosen subset of the dataset. This means that the models are predicting more unseen data than seen data in hopes of showing the ability of the models for a limited sample size.

\section{Results}
\begin{figure}
    \centering
    \includegraphics[width=\linewidth]{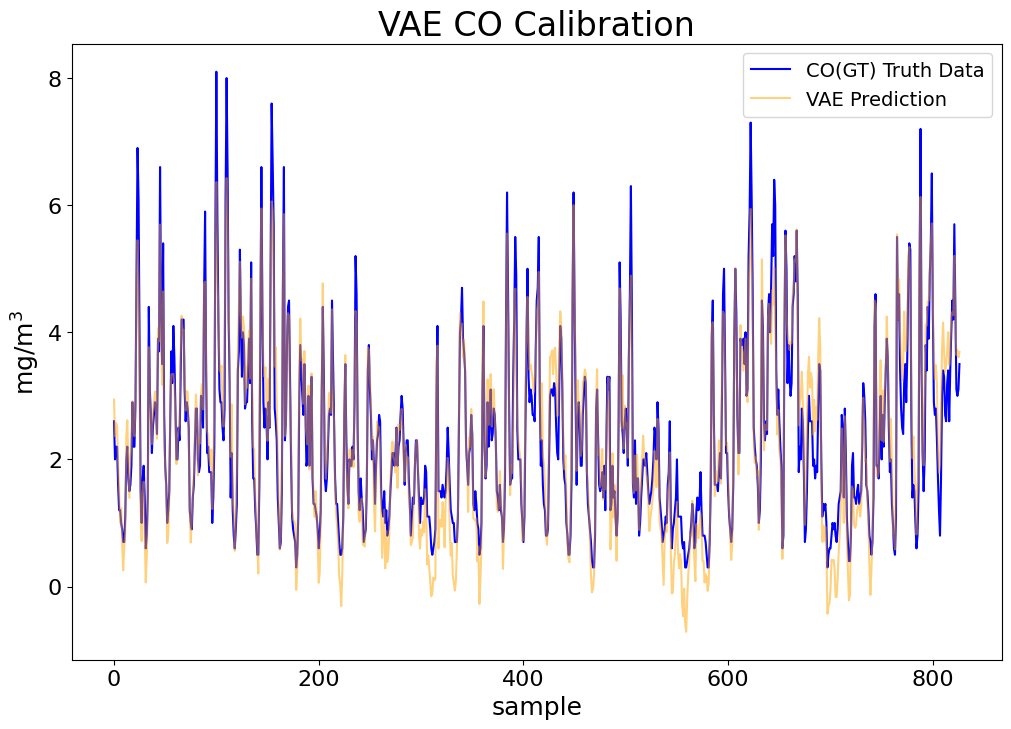}
    \caption{\textcolor{black}{Plot showing the predictions of the CO VAE and truth data from CO(GT) sensor as seen in Table \ref{tab:CO_features_metrics_summary}.}}
    \label{fig:CO_predict}
\end{figure}

\begin{figure}
    \centering
    \includegraphics[width=\linewidth]{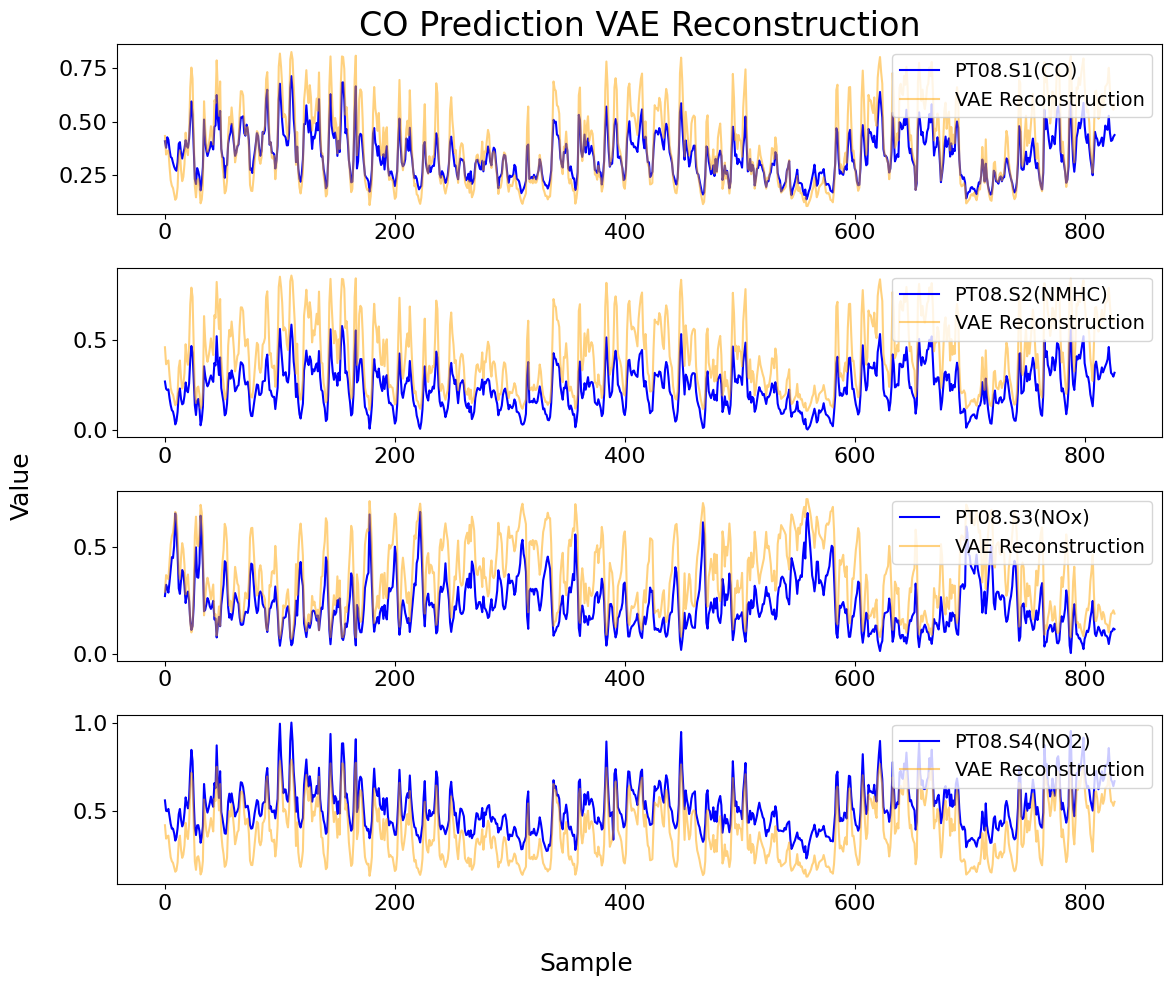}
    \caption{\textcolor{black}{Plot showing the reconstruction of the input sensor data produced by the CO VAE as seen in Table \ref{tab:CO_features_metrics_summary}.}}
    \label{fig:CO_reconstruct}
\end{figure}

\begin{figure}
    \centering
    \includegraphics[width=\linewidth]{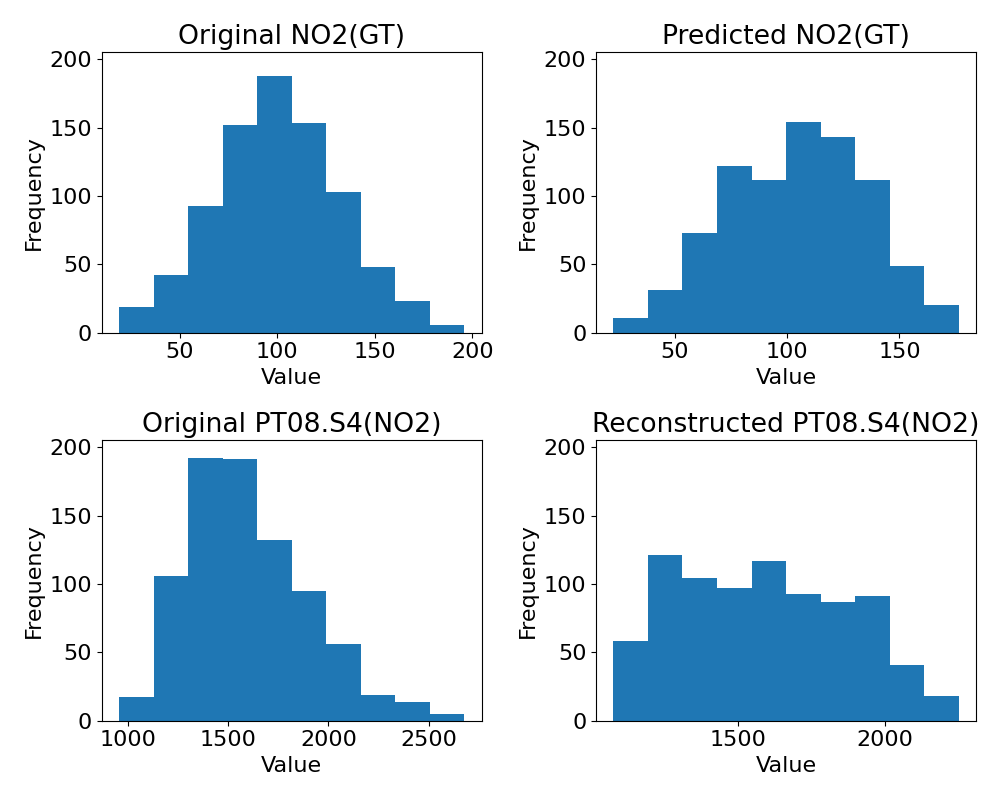}
    \caption{Histogram of 4 separate components from the NO$_2$ model test. In the top row we see the truth data ($Y$) distribution from the "NO2(GT)" sensor and the model predictions ($Y'$) from the latent space. In the bottom row we see the "PT08.S4(NO2)" MOS input data (a component of $X$) distribution as well as the reconstruction data (a component of $X'$) distribution produced by the model. These symbols are in reference to Figure \ref{fig:Vae_prop}.}
    \label{fig:NO2_distribution}
\end{figure}

This section contains the results of the models trained to calibrate each of the four chosen ground truth sensor gases. Table \ref{tab:CO_features_metrics_summary} shows the performance of the model trained to predict the GT(CO) matched values as well as the scores of the model's ability to reconstruct the inputs \textcolor{black}{from} the latent vector. The same is done for the remaining target sensors in Tables \ref{tab:NMHC_features_metrics_summary}, \ref{tab:NOx_features_metrics_summary} and \ref{tab:NO2_features_metrics_summary}. \textcolor{black}{Figures \ref{fig:CO_predict} and \ref{fig:CO_reconstruct} show the predictions of the calibration output and the reconstruction of the CO calibration VAE presented in Table \ref{tab:CO_features_metrics_summary}}. Figure \ref{fig:NO2_distribution} shows the input and output distributions of the NO$_2$ calibration VAE.

\begin{table}[]
\centering
\caption{Table showing reconstruction and calibration of CO performance of the VAE over test data}
\resizebox{0.48\textwidth}{!}{%
\begin{tabular}{|l|l|l|l|l|l|l|l|}
\hline
\textbf{Sensor}                 & \textbf{MAE} & \textbf{Accuracy (\%)} & \textbf{R\textsuperscript{2}} & \textbf{KL Divergence} & \textbf{JS Divergence} \\ \hline
\textbf{PT08.S1(CO)} & 64.28         & 94.55         & 0.8940        & 0.00998       & 0.00250       \\
\textbf{PT08.S2(NMHC)} & 39.13         & 95.44         & 0.9656        & 0.00885       & 0.00222       \\
\textbf{PT08.S3(NOx)} & 92.65         & 90.39         & 0.8212        & 0.02322       & 0.00579       \\
\textbf{PT08.S4(NO$_2$)} & 46.98         & 96.98         & 0.9561        & 0.01995       & 0.00500       \\ \hline
\textbf{CO(GT)} & 0.29         & 83.12         & 0.9319        & 0.01309      & 0.00326       \\ \hline
\end{tabular}
\label{tab:CO_features_metrics_summary}
}
\end{table}

\begin{table}[]
\centering
\caption{Table showing reconstruction and calibration of NMHC performance of the VAE over test data}
\resizebox{0.48\textwidth}{!}{%
\begin{tabular}{|l|l|l|l|l|l|l|l|}
\hline
\textbf{Sensor}                 & \textbf{MAE} & \textbf{Accuracy (\%)} & \textbf{R\textsuperscript{2}} & \textbf{KL Divergence} & \textbf{JS Divergence} \\ \hline
\textbf{PT08.S1(CO)} & 82.92 & 93.02 & 0.8288 & 0.0086 & 0.00215       \\
\textbf{PT08.S2(NMHC)} & 47.68 & 94.66 & 0.9504 & 0.00604 & 0.00151       \\
\textbf{PT08.S3(NOx)} & 67.83 & 93.05 & 0.8963 & 0.01501 & 0.00374     \\
\textbf{PT08.S4(NO$_2$)} & 49.39 & 96.91 & 0.9522 & 0.01434 & 0.00357
       \\ \hline
\textbf{NMHC(GT)} &79.25 & 43.19 & 0.6596 & 0.10152 & 0.02451     \\ \hline
\end{tabular}
\label{tab:NMHC_features_metrics_summary}
}
\end{table}

\begin{table}[]
\centering
\caption{Table showing reconstruction and calibration of NOx performance of the VAE over test data}
\resizebox{0.48\textwidth}{!}{%
\begin{tabular}{|l|l|l|l|l|l|l|l|}
\hline
\textbf{Sensor}                 & \textbf{MAE} & \textbf{Accuracy (\%)} & \textbf{R\textsuperscript{2}} & \textbf{KL Divergence} & \textbf{JS Divergence} \\ \hline
\textbf{PT08.S1(CO)} & 62.65 & 94.74 & 0.8988 & 0.01048 & 0.00261       \\
\textbf{PT08.S2(NMHC)} & 48.84 & 94.43 & 0.9466 & 0.00743 & 0.00186  \\
\textbf{PT08.S3(NOx)} &  93.44 & 90.29 & 0.8205 & 0.02069 & 0.00516      \\
\textbf{PT08.S4(NO$_2$)} & 38.83 & 97.56 & 0.9664 & 0.01641 & 0.00410
       \\ \hline
\textbf{NOx(GT)} & 26.76 & 73.21 & 0.8284 & 0.03862 & 0.00958      \\ \hline
\end{tabular}
\label{tab:NOx_features_metrics_summary}
}
\end{table}

\begin{table}[]
\centering
\caption{Table showing reconstruction and calibration of NO$_2$ performance of the VAE over test data}
\resizebox{0.48\textwidth}{!}{%
\begin{tabular}{|l|l|l|l|l|l|l|l|}
\hline
\textbf{Sensor}                 & \textbf{MAE} & \textbf{Accuracy (\%)} & \textbf{R\textsuperscript{2}} & \textbf{KL Divergence} & \textbf{JS Divergence} \\ \hline
\textbf{PT08.S1(CO)} & 59.49 & 95.07 & 0.9018 & 0.01136 & 0.00284       \\
\textbf{PT08.S2(NMHC)} & 42.4 & 95.52 & 0.9567 & 0.00749 & 0.00187  \\
\textbf{PT08.S3(NOx)} &  91.0 & 90.08 & 0.845 & 0.01221 & 0.00304      \\
\textbf{PT08.S4(NO$_2$)} & 58.09 & 96.43 & 0.9242 & 0.01695 & 0.00421
       \\ \hline
\textbf{NO$_2$(GT)} & 11.7 & 86.68 & 0.7606 & 0.01041 & 0.00260    \\ \hline
\end{tabular}
\label{tab:NO2_features_metrics_summary}
}
\end{table}

\section{Discussion}
As we can see from the results obtained in Tables \ref{tab:CO_features_metrics_summary}-\ref{tab:NO2_features_metrics_summary}, the statistical properties of this data show a great similarity to the original data for both the inputs and the calibration. The worst performer for calibration was the NMHC calibration, seen in Table \ref{tab:NMHC_features_metrics_summary}, reporting an order of magnitude higher error for the Kullback-Leibler and Jensen-Shannon divergences when compared to that of the other models. It was also the lowest scoring calibration model across the R$^2$ and percentage accuracy indicating its calibration capability was not nearly as successful as the other implementations on this data. It is interesting to note that all the models were very accurate when recreating the input data. This indicates that \textcolor{black}{the} proposed VAE was functioning as a standard VAE with the added condition of providing a calibration output from the latent space, which in this instance is a one-dimensional output. This shows that when the latent space is trained as a calibration output, the model is able to encode from uncalibrated inputs to the expected calibrated output as well as from the calibrated latent vector back to the expected inputs.  

\textcolor{black}{Looking at Figures \ref{fig:CO_predict} and \ref{fig:CO_reconstruct} we see the prediction and reconstruction of the CO prediction VAE respectively. These are the results that relate to Table \ref{tab:CO_features_metrics_summary} showing the actual performance of the model visually. As can be seen, both the predictions and the reconstructions are highly similar to their target sensor data and show that the model is capable of approximating both the calibration data and the reconstruction data simultaneously. Looking that the work done by Yu et al. \cite{yu2020deep}, they show an implementation of their calibration model being tested on the same dataset as used in our work. It is important to note that the models are very different and they were able to use more of the dataset because of this. However, it is encouraging to note that the MAE calibration performance that our model achieved of $0.29$ , seen in Table \ref{tab:CO_features_metrics_summary}, is almost on par with their DeepCM model's MAE of $0.288$ which was the best performer of the methods they tested on this dataset. This shows the performance of the calibration is similar to existing methods while performing the VAE reconstruction simultaneously.}

A sample of the probability distribution of the inputs and outputs \textcolor{black}{produced by} the NO$_2$ model experiment  can be seen in Figure \ref{fig:NO2_distribution}. We can see the probability distributions show some interesting differences. Looking at Figure \ref{fig:NO2_distribution} with reference to Figure \ref{fig:Vae_prop}, when comparing the NO2(GT) which is the $Y$ to the portion of the input $X$ represented by the PT08.S4(NO2), we see that the distributions are not the same. It is clear that the MOS sensor does not produce a Gaussian distribution but rather a distribution more similar to a Rayleigh distribution. We also find that both the $Y'$, represented by the predicted NO2(GT), and the portion of the reconstruction $X'$, represented by reconstruction PT08.S4(NO2), show differences to their counterparts. Specifically they have a narrower \textcolor{black}{range} of values and appear more uniformly distributed, however they both fall within the bounds of their counterparts.

\section{Conclusion}
We have shown a working, new, proposed implementation of a Variational Autoencoder for sensor calibration. \textcolor{black}{It show promise calibrating sensor data using a group of cross-sensitive sensors as an input.} This shows the ability for the latent space of the VAE to be used as a calibration output and to capture the properties of the sensor within the VAE. Specifically, these models have shown that they are capable of creating statistically similar outputs from the calibration output as well as the reconstruction output. We believe this indicates that the models are capturing the properties of the transfer functions between the environment, the sensor and the calibrated truth data.
With these findings we are encouraged to continue to investigate the uses of this architecture and to evaluate the performance of different loss functions for training. This may include further statistical analysis on the input data to find more effective losses for non-Gaussian data. We wish to apply this model to new datasets to further understand its capability as a calibration model, such as the dataset developed in our previous work \cite{barrett2024statistical}.


\section*{Acknowledgment}
We would like to thank Sentech Soc Ltd for their assistance with our research. We would also like to thank the European Commission Erasmus+ program.

\bibliographystyle{IEEEtran}
\bibliography{Bib}

\end{document}